\documentclass[fleqn]{NCMAI2026}
\usepackage{titlesec}
\usepackage{hyperref}
\usepackage{graphicx} 
\usepackage{multirow}
\usepackage{booktabs}
\usepackage{pifont}
\usepackage{xcolor}
\usepackage{array}
\usepackage{makecell}
\usepackage{natbib}
\usepackage{amssymb}
\usepackage{multicol}
\usepackage[T1]{fontenc} 

\newcolumntype{R}[1]{>{\raggedleft\arraybackslash}m{#1}}
\newcolumntype{L}[1]{>{\raggedright\arraybackslash}m{#1}}
\newcolumntype{C}[1]{>{\centering\arraybackslash}m{#1}}

\usepackage{microtype}

\newcommand{\cmark}{\textcolor{green!60!black}{\ding{51}}}   
\newcommand{\pmark}{\textcolor{orange!80!black}{$\sim$}}      
\newcommand{\xmark}{\textcolor{red!70!black}{\ding{55}}}      

\begin{document}

\title{Automated Scoring of Arabic Text Using Large Language Models: A Literature Review}

\firstpagehead{}
\runningheads{The First National Conference
	on Mathematics and Artificial
	Intelligence (NCMAI'26)}{}
\runnigfoots{}

\author{Khaoula Dahimi$^{1,*}$, Hadda Cherroun$^{1}$, Amel Belabbaci$^{1}$}

\address{
Laboratoire d'Informatique et de Mathématiques LIM\\
\textit{Amar Telidji University},
Laghouat, Algeria \\
*Corresponding author: \href{k.dahimi@lagh-univ.dz}{k.dahimi@lagh-univ.dz}
}

\fontfamily{phv}\fontseries{mc}\fontsize{9}{10}\selectfont
\linespread{1.2}
\vskip 8 mm
\abstract{
In modern educational systems, Automatic Text Scoring (ATS) plays a central role by enabling scalable and consistent evaluation of learner responses without human intervention. Recently, the increased accessibility of LLMs and Arabic-specific datasets has sparked renewed interest in this area. In this work, we investigate LLM-Based approaches for the automated evaluation of Arabic texts, focusing on both short answer grading (ASAG) and essay scoring (AES). We further introduce a structured taxonomy comprising five dimensions: application domain, feedback generation capability, LLM architecture deployed, alignment with competency referential frameworks, and prompt engineering strategy. By applying this taxonomy, we conduct a comparative analysis of existing studies, examining their methodological approaches, datasets, evaluation metrics, and reported performance. The findings highlight the need for sustained and pedagogically grounded research efforts in Arabic ATS, given its significance for improving educational quality across Arabic-speaking  communities.
}

\section{Introduction}
\label{Sec:Intro}

 Automatic Scoring of Text (ATS) uses Natural Language Processing (NLP) to grade written answers with no human grader needed. Early ATS systems relied on carefully selected linguistic features and traditional statistical models to evaluate grammar, vocabulary, coherence, and whether the response actually answered the question. But Deep Learning changed the game. With transformer-based models, these systems became significantly smarter. Now, they are able to understand the context and meaning behind the words~\citep{alikaniotis2016automatic}.

Then came large language models (LLMs), and things moved even faster. Instead of just counting features or scoring surface errors, LLMs use huge pretraining datasets and advanced language modeling. That means they can handle difficult content like discourse coherence, sentence structure, meaning, and even style~\citep{alikaniotis2016automatic}. Moreover, they don't simply generate a score they are able to provide meaningful feedback, which creates new opportunities for teaching and learning~\citep{fagbohun2024beyond}.

While LLM-based scoring has gained increasing attention in high-resource languages,  research on Arabic is currently limited and dispersed. Arabic throws in extra challenges complex morphology, inconsistent orthography, and all those dialects which may challenge automated systems~\citep{nael2022arascore}. Moreover, existing studies vary significantly in modeling strategies, prompt design, evaluation criteria, and reported outcomes, making systematic comparison difficult. Our Arabic ATS literature focuses on this gap.

This paper presents a structured and comprehensive overview of ATS with a particular focus on Arabic. It systematically reviews existing approaches that integrate LLMs into ATS frameworks and critically examines their limitations. Furthermore, it compares prior work across five key dimensions: application domain, employed LLMs, competency frameworks, prompting strategies, and feedback generation. Finally, it identifies major research gaps and outlines future directions toward the development of more robust, reliable, and standardized Arabic ATS systems.

The remainder of this paper is organized as follows. In Section~\ref{sec:background} we introduce  domain preliminaries. Section~\ref{sec:methodology} presents our reviewing  methodology and taxonomy. Section~\ref{Sec:Survey} is followed by  a comparative analysis of LLM-based approaches for automated Arabic text scoring. Section~\ref{sec:RG} discusses the key limitations and research gaps identified in the existing literature. Finally, Section~\ref{sec:conclusion} concludes the paper.

\section{Background}
\label{sec:background}
In educational contexts, assessment systems play a vital role in grading student responses, whether short answers or essays to enhance learning outcomes while saving teachers' time and effort.

Short answer scoring (ASAG) and essay scoring (AES) differ primarily in response length, evaluation focus, and techniques used in automated systems. Table~\ref{tab:asag_vs_aes} summarizes these distinctions.

\begin{table}[h!]
\centering\small
  \renewcommand{\arraystretch}{1}
\begin{tabular}{L{2.2cm} L{4cm} L{4cm}}
\toprule
\textbf{Aspect} & \textbf{Short Answer Scoring (ASAG)} & \textbf{Essay Scoring (AES)} \\ \midrule
Text Length & Brief (sentences/paragraphs)  & Extended (multi-paragraph) \\ \midrule
Key Focus & Content accuracy, keyword overlap  & Coherence, style, development \\ \midrule

Challenges & Domain-specific meaning & Subjectivity in creativity  \\ \bottomrule

\end{tabular}
\caption{Comparison between Short Answer Scoring (ASAG) and Essay Scoring (AES)}
\normalsize 
\label{tab:asag_vs_aes}
\end{table}

\subsection*{Evaluation metrics}
 To evaluate ATS systems, researchers rely on well-established metrics that assess both the quality and accuracy of learners’ answers~\citep{ramesh2022automated,alqurashi2025automatic}.
 
 When ATS is treated as an ordinal classification task, metrics such as \textbf{Accuracy}, \textbf{F1-score}, and \textbf{Root Mean Squared Error (RMSE) }are commonly used. These metrics quantify either the exact match between predicted and actual scores or the average magnitude of prediction errors.

\textbf{Quadratic Weighted Kappa (QWK)} is a widely adopted metric, as it measures the level of agreement between two independent raters while accounting for the degree of disagreement~\citep{doewes2023evaluating}. Unlike simple accuracy, QWK considers how far apart the predicted and true scores are, giving more credit to near matches than to large discrepancies. It is particularly favored because it adjusts for chance agreement and emphasizes the ordinal nature of scoring~\citep{ghazawi2025well}.


\section{Methodology and Taxonomy}
\label{sec:methodology}
 Recent advances in Arabic ATS reflect a clear methodological shift from traditional approaches toward Deep Learning and, more recently, large language models. While earlier work relied on handcrafted linguistic features combined with classical machine learning pipelines \citep{lotfy2023enhanced}, the present review focuses exclusively on Deep Learning-based systems, with particular emphasis on LLM-driven approaches. The scope is deliberately narrowed to the period  [2020,~2025]. This period is intentionally chosen as it witnessed both the emergence and rapid expansion of LLMs usage in NLP tasks. 
 
 To structure our analysis, we propose a taxonomy organized around five criteria: 
 
 \begin{enumerate}
 \item  Domain of application: Educational domains differ in their linguistic and cognitive demands, so documenting this dimension reveals whether a model truly generalizes or merely overfits to a narrow subject area. It also maps the coverage of existing research, exposing underserved domains that still lack Arabic ATS solutions.
 \item  Feedback generation capability: Meaningful feedback is a cornerstone of formative assessment, yet most systems only output a score. Tracking this criterion exposes a critical gap, since LLMs are inherently capable of generating natural language explanations but this potential remains largely unexploited.
 \item LLM architecture deployed:The choice between a fine-tuned encoder like AraBERT and a generative model like GPT-4 determines how well a system handles Arabic's morphological complexity and contextual nuance. Comparing architectures helps identify which model families are best suited for Arabic ATS under different conditions.
 
 \item Alignment with competency referential frameworks: A scoring system disconnected from recognized educational standards produces results that are technically computed but pedagogically meaningless. This criterion reveals whether automated scores actually reflect the normative expectations that govern real classroom evaluation.
 
\item Prompt engineering deployed strategies: For generative LLMs, prompt design directly and measurably affects scoring accuracy and consistency. Cataloging these strategies allows the community to identify best practices and highlights that current approaches remain largely ad hoc and lack principled pedagogical grounding.
\end{enumerate}

\begin{figure}
    \centering
    \includegraphics[width=\linewidth]{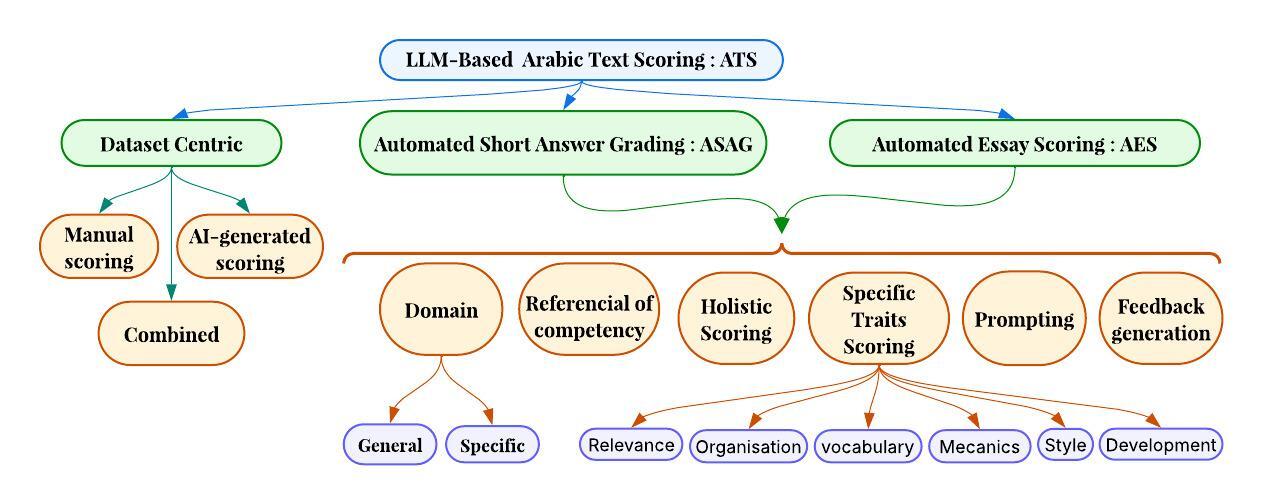}
    \caption{The Proposed Taxonomy for ATS frameworks.}
    \label{fig:Taxonomy}
\end{figure}

\section{ATS Related  Datasets and Approaches}
\label{Sec:Survey}
Figure~\ref{fig:Taxonomy} illustrates the proposed taxonomy  according to the  previously defined criteria.  In fact, the papers selected under this review  are examined in detail in the following sections according to the three main  categories: \textit{Dataset Centric} investigations,  \textit{Automated Short Answer Scoring} and \textit{Automated Essay Scoring} Approaches.

\subsection{Dataset Centric}
\label{Sec:DataCentric}
Due to the scarcity of datasets for  ATS in the Arabic language, most pioneer studies have started by building dedicated datasets. The general form of an instance consists of a prompt paired with the learner's response and one or more holistic scores, depending on the number of graders (assigned either manually or automatically), along with optional Trait-Specific Scoring. By prompt, we refer to the task-specific instruction, assignment, or writing topic that the examinee is asked to respond to. Trait-Specific Scoring refers to a set of detailed scores evaluated along specific dimensions of the response, such as relevance, organization, vocabulary, style, development, mechanics, and grammar. 

Additional annotations are  included such  morphological segmentation, part-of-speech tagging, lemmatization, and  CEFR\footnote{\tiny CEFR :Common European Framework of Reference for Languages, a standard developed  to describe and measure language proficiency across six levels: A1, A2 , B1, B2, C1, and C2.} proficiency levels ranging from A2 to C1. 
Table~\ref{tab:datasets} summarizes the landscape of publicly available datasets for Arabic ATS.

    
ATS datasets remains relatively limited in domain, scale, and annotation depth. Most of the identified datasets rely on manual annotation, which, while ensuring quality, limits scalability. Examples of fully manually annotated resources include \emph{ZAEBUC}~\citep{habash2022zaebuc}, \emph{QAES}~\citep{bashendy2024qaes} (notably the first publicly available Arabic essay scoring dataset), and Sociology-SAS~\citep{shehab2026arabic}.

In addition, we observe that the sizes of datasets  vary considerably, with LAILA~\citep{bashendy2025laila} which is the largest publicly available dataset   appearing as a notably larger collection, whereas others appear to be smaller in scale such ZAEBUC and Sociology-SAS.

Regarding annotation depth, most AES datasets stand out by providing  trait-specific scores, as reflected in the corresponding column. This dual scoring approach enhances evaluation quality and supports more informative feedback generation. In contrast, all ASAG datasets offer only a single overall score, limiting their capacity for fine-grained diagnostic assessment.

\begin{table}[h]
\centering
\caption{Comparison of Arabic ATS Datasets}
\resizebox{\textwidth}{!}{%
\begin{tabular}{lllrcccc}
\hline
\textbf{Dataset} & \textbf{Language} & \textbf{Annotation} & \textbf{Size} & \textbf{\# Prompts} &  \textbf{Trait-Specific Score} & \textbf{Linguistic Annotation} & \textbf{CEFR} \\
\hline
\multicolumn{8}{c}{ Arabic Essay Scoring}\\\hline
\emph{ZAEBUC}~\citep{habash2022zaebuc} & Bilingual (AR--EN) & Manual & 214 essays & 3 & \texttimes & \checkmark (POS, morph, lemma) & A2--C1 \\
\emph{QAES}~\citep{bashendy2024qaes}   & Arabic           & Manual & 195 essays & 2 &  \checkmark (7 traits) & \texttimes & \texttimes \\
\emph{ZaQQ}~\citep{elsayed2025zaqq} & Arabic & Manual \& Automatic & 1021 essays & 5 & \checkmark & \texttimes  & \texttimes \\
\emph{AR-AES}~\citep{Ghazawi_Simpson_2024}   & Arabic           & Manual & 2046 essays & 12 &  \texttimes  & \texttimes & \texttimes \\
\emph{TAQEEM}~\citep{bashendy2025taqeem} & Arabic           & Manual & 1,265 essays & Multiple & \checkmark & \texttimes & \texttimes \\
\emph{LAILA}~\citep{bashendy2025laila}  & Arabic           & Manual & 7,859 essays & 8        &  \checkmark & \texttimes & \texttimes \\\hline
\multicolumn{8}{c}{ Arabic Short Answer Scoring}\\\hline
\emph{ASAP-SAS} (Translated)\citep{nael2022arascore} & English/Arabic & Manual & 17,205 answers & 10 &   \texttimes & \texttimes &  \texttimes  \\
\emph{AR-ASAG}~\citep{ouahrani2020ar} &  Arabic   & Manual & 2,133 answers & $48$  &  \texttimes & \texttimes & \texttimes  \\
\emph{Sociology SAS}~\citep{shehab2026arabic}& Arabic & Manual & 270 answers& $27$ & \texttimes & \texttimes& \texttimes\\
\emph{ESD}~\citep{nabil2025leveraging} & Arabic & Manual & 610 answers& 61&  \texttimes & \texttimes & \texttimes \\
\hline
\end{tabular}%
}
\label{tab:datasets}
\end{table}

Overall, ATS faced significant challenge due to data scarcity, as publicly available annotated datasets are limited and considerably smaller than their English counterparts~\citep{bashendy2025laila}. This restricts the generalization of the built models. In addition, this  fact has  lead many studies to rely on collecting domain-specific customized datasets.

\subsection{Automated Short Answer Grading}
\label{Sec:ASAG}
Existing Arabic ASAG approaches differ in the methodologies they employ. Studies~\citep{nael2022arascore,soulimani2024deep,noaman2025leveraging} proposed transformer-based systems that leverage transfer learning for educational contexts. In contrast, \citet{Badry_Ali_Rslan_Kaseb_2023} introduced a local weight-based LSA model, which demonstrated strong performance in measuring semantic overlap. Additionally, \citet{nabil2025leveraging} evaluated LLMs and their capacity to generate feedback using an Environmental Science dataset. Table~\ref{tab:asag} summarizes these approaches according to the taxonomy defined in Section~\ref{sec:methodology}.

\begin{table}[htbp]
\centering
\caption{Comparison of LLM-Based Arabic Automated \textbf{Short Answer Grading} (ASAG) Approaches}
\label{tab:asag}
\small
\setlength{\tabcolsep}{3pt}
\renewcommand{\arraystretch}{1.1}
\resizebox{\linewidth}{!}{%
 \begin{tabular}{
  L{2cm}         
  L{3.2cm}         
  C{2cm}         
  L{3.2cm}         
  C{2cm}         
  C{2cm}         
  C{2cm}         
}
\toprule
\textbf{Approach} &
\makecell[l]{\textbf{Application}\\\textbf{Domain}} &
\makecell{\textbf{Feedback}\\\textbf{Generation}} &
\makecell[l]{\textbf{LLM /}\\\textbf{Model}} &
\makecell{\textbf{Competency}\\\textbf{Framework}} &
\makecell{\textbf{Prompt}\\\textbf{Engineering}} &
\textbf{Score} \\
\midrule

\citet{nael2022arascore}
  & General\newline\footnotesize(ASAP-SAS)
  & \xmark
  & AraBERT, ELECTRA, Bi-LSTM, RNN
  & \xmark
  & \xmark
  & QWK = 0.78 \\
\midrule
\citet{Badry_Ali_Rslan_Kaseb_2023}
  & Cybercrime Course\newline\footnotesize(AR-ASAG)
  & \xmark
  & Local weight-based LSA
  & \xmark
  & \xmark
  & F1 = 82.82\% \\
\midrule
\citet{soulimani2024deep}
  & Islamic Edu.\newline\footnotesize(Moroccan pupils)
  & \xmark
  & AraBERT, LSTM
  & \xmark
  & \xmark
  & Acc = 71.31\% \\
\midrule
\citet{noaman2025leveraging}
  & Cybercrime Course\newline\footnotesize(AR-ASAG)
  & \xmark
  & AraBERT (Siamese network)
  & \xmark
  & \xmark
  & Pearson = 0.79 \\
\midrule
\citet{nabil2025leveraging}
  & Science\newline\footnotesize(ESD)
  & \cmark
  & GPT-4, Llama~3, DeepSeek-V3
  & \xmark
  & \cmark
  & QWK = 0.83 \\

\bottomrule
\end{tabular}
}
\smallskip
\begin{minipage}{\linewidth}
\ \centering
\footnotesize
\cmark\;=\;present;\quad
\xmark\;=\;absent.
\medskip\par
QWK~=~Quadratic Weighted Kappa;\quad F1~=~F1-Score;\quad Acc~=~Accuracy;\quad Pearson~=~Pearson correlation coefficient.  
\end{minipage}
\end{table}
The aforementioned studies exhibit several recurring limitations rooted in their underlying frameworks. First, a number of studies (\citep{nael2022arascore,soulimani2024deep,nabil2025leveraging}) rely on non-authentic or constrained datasets, such as translated benchmarks or small, domain-specific corpora, which limits linguistic validity, generalizability, and overall robustness. Second, other approaches (\citet{Badry_Ali_Rslan_Kaseb_2023,noaman2025leveraging}) demonstrate an over-reliance on semantic similarity and lexical overlap, restricting their ability to capture deeper reasoning and partial understanding, and making them less effective when evaluating correctly answered responses expressed in different forms. Finally, the variation in datasets and evaluation metrics across these studies makes it difficult to determine which approach achieves state-of-the-art performance.

\subsection{Automated Essay Scoring}
\label{Sec:AES}
For AES, only few studies have been conducted in  Arabic language. Table~\ref{tab:comparison_AES}  summarizes these studies according to our proposed taxonomy defined in Section~\ref{sec:methodology}. 

Unlike the ASAG approaches, the existing studies shift toward zero-shot and few-shot prompting using multilingual and Arabic-focused LLMs.  Studies~\citep{ghazawi2025well,bashendy2025laila}  performed  comparative evaluation using a set of LLMs across various evaluation methodologies, including zero-shot learning and prompt engineering strategies. Additionally, \citet{almarwani2025taibah,Alnajjar_2025} introduced a grading rubric system based on GPT-4 model. 

\begin{table}[htbp]
\centering
\caption{Comparison of LLM-Based Arabic Automated \textbf{Essay Scoring} (AES) Approaches}
\label{tab:comparison_AES}
\small
\setlength{\tabcolsep}{5pt}
\renewcommand{\arraystretch}{1.45}
 \resizebox{\linewidth}{!}{%
\begin{tabular}{
  L{2.0cm}         
  L{3.5cm}         
  C{2cm}         
  L{3.0cm}         
  C{2.2cm}         
  L{2.4cm}         
  C{1.4cm}         
}
\toprule
\textbf{Approach} &
\makecell[l]{\textbf{Application}\\\textbf{Domain / Used Dataset }}   &
\makecell{\textbf{Feedback}\\\textbf{Generation}} &
\makecell[l]{\textbf{LLM /}\\\textbf{Model}} &
\makecell{\textbf{Competency}\\\textbf{Framework}} &
\makecell[l]{\textbf{Prompt}\\\textbf{Engineering}} &
\textbf{Score (QWK)} \\
\midrule

\citet{Ghazawi_Simpson_2024}
  & Info. Science, Chemistry, Biotechnology (AR-AES)
  & \xmark
  & AraBERT (fine-tuned)
  & \xmark
  & \xmark
  & $ 0.88$ \\
\midrule

\citet{ghazawi2025well}
  & Info. Science, Chemistry, Biotechnology (AR-AES)
  & \xmark
  & ChatGPT, Llama, Aya, Jais, ACEGPT
  & \xmark
  & \pmark\;Mixed-language prompting
  & $0.67$ \\
\midrule

\citet{almarwani2025taibah}
  & General\newline(TAQEEM)
  & \xmark
  & GPT-4o
  & \xmark
  & \cmark\;Few-shot + grading rubric
  & $0.65$ \\
\midrule

\citet{Alnajjar_2025}
  & General\newline(TAQEEM)
  & \xmark
  & GPT-4.1
  & \xmark
  & \cmark\;Few-shot + grading rubric
  & $0.612$ \\
\midrule

\citet{bashendy2025laila} 
  & General\newline(LAILA)
  & \xmark
  & Fanar, ALLaM, R7B Arabic
  & \xmark
  & \pmark\;Prompt-specific strategies
  & $0.58$ \\

\bottomrule
\end{tabular}
}
\smallskip
\begin{minipage}{\linewidth}
\footnotesize\centering
\cmark\;=\;present;\quad
\pmark\;=\;partial / not fully reported;\quad
\xmark\;=\;absent.
\medskip\par
QWK~=~Quadratic Weighted Kappa.
\end{minipage}
\end{table}

These studies highlight several important trends and limitations. First, earlier work focuses on domain-constrained datasets, whereas more recent studies adopt general-domain datasets such as TAQEEM and LAILA. This shift suggests a growing interest in building more robust and generalizable AES systems. Second, the choice of LLMs demonstrates a clear evolution, with recent approaches leveraging advanced generative models. Despite this shift, performance gains remain inconsistently significant. Finally, with respect to feedback generation, all approaches lack this capability. This constitutes a notable gap, as modern educational systems increasingly require not only scores but also formative, interpretable feedback to support learning.

\section{Discussion}
\label{sec:RG}
The study of these approaches demonstrates that the field has attracted few attention despite those  commendable efforts that  have been made toward automating ATS. The diversity of models explored from fine-tuned transformer architectures such as AraBERT and ELECTRA to frontier large language models including GPT-4, DeepSeek-V3, and Llama 3, alongside the development of dedicated Arabic datasets, reflects a growing and dynamic research community. However, a closer examination of the reviewed work reveals several  gaps that the field has yet to address.

First, none of the examined approaches addresses the scoring of visually-grounded assignments (such as diagrams, annotated figures, or multimodal responses) despite the growing prevalence of such tasks in modern educational contexts.

Second, feedback generation remains largely absent across both ASAG and AES systems. With the single exception of Nabil et al. (2025), none of the reviewed work produces explanatory feedback for learners. Yet meaningful feedback is not merely a supplementary feature, it is a cornerstone of formative assessment and a prerequisite for effective learning. The capacity of LLMs to generate natural language explanations makes this gap particularly easy to resolve.

Third, while prompt engineering has emerged as a promising strategy for LLM-based scoring, current approaches lack grounding in established educational standards. No study aligns its prompting strategy with a recognized competency referential framework, such as national curricula, learning outcome taxonomies, or subject-specific rubrics. Without such grounding, scores produced by these systems remain disconnected from the normative expectations that govern real educational evaluation.

Fourth, almost reviewed approaches  treat scoring as a static, one-shot prediction task. None integrates of more  learnable mechanism capable of adapting to the specific content and objectives of a given assignment. Incorporating the assignment text itself alongside its target competencies and marking scheme, as explicit input to the model would enable more contextualized, assignment, aware scoring and open the door to systems that improve over time through accumulated grading experience.

\section{Conclusion}
\label{sec:conclusion}

We have examined automated ATS approaches based on Deep Learning and LLMs, revealing both encouraging progress and persistent gaps. While commendable efforts have been made, critical dimensions such as feedback generation, alignment with competency frameworks, marking scheme integration, and visual assignment scoring remain largely unaddressed. 

Closing the identified gaps represents not only a technical challenge but a meaningful contribution to educational equity and learning quality. This domain deserves sustained research attention.

\bibliographystyle{plainnat}  
\bibliography{references}

\end{document}